\documentclass[conference]{IEEEtran}
\IEEEoverridecommandlockouts
\usepackage{cite}
\usepackage{amsmath,amssymb,amsfonts}
\usepackage{algorithm}
\usepackage{algorithmicx}
\usepackage{graphicx}
\usepackage{textcomp}
\usepackage{xcolor}
\usepackage{subfigure}
\usepackage{setspace}
\usepackage{algpseudocode}
\usepackage{longtable}
\usepackage{multirow}
\usepackage{authblk}

\usepackage{color}

\def\BibTeX{{\rm B\kern-.05em{\sc i\kern-.025em b}\kern-.08em
    T\kern-.1667em\lower.7ex\hbox{E}\kern-.125emX}}
\begin{document}

\title{Fast Local Attack: Generating Local Adversarial Examples for Object Detectors
\thanks{$^{*}$Xin Wang~(xinw@curacloudcorp.com) and Xi Wu~(xi.wu@cuit.edu.cn) are corresponding authors.}
}


\author[1]{Quanyu Liao}
\author[2]{Xin Wang$^*$ \IEEEmembership{Member,~IEEE}} 
\author[2]{\\Bin Kong}
\author[3]{Siwei Lyu \IEEEmembership{Senior Member,~IEEE}}
\author[2]{Youbing Yin}
\author[2]{Qi Song}
\author[1]{Xi Wu$^*$}
\affil[1]{Chengdu University of Information Technology, Chengdu, China}
\affil[2]{CuraCloud Corporation, Seattle, USA}
\affil[3]{SUNY Albany, NY, USA}


\maketitle

\begin{abstract}
The deep neural network is vulnerable to adversarial examples. Adding imperceptible adversarial perturbations to images is enough to make them fail. Most existing research focuses on attacking image classifiers or anchor-based object detectors, but they generate globally perturbation on the whole image, which is unnecessary. 
In our work, we leverage higher-level semantic information to generate high aggressive local perturbations for anchor-free object detectors. As a result, it is less computationally intensive and achieves a higher black-box attack as well as transferring attack performance. The adversarial examples generated by our method are not only capable of attacking anchor-free object detectors, but also able to be transferred to attack anchor-based object detector.
\end{abstract}


\section{Introduction}
The development of deep neural networks (DNNs) supports researchers to achieve unprecedented high performance in various computer vision problems. Nevertheless, these deep learning-based algorithms are notoriously vulnerable to adversarial examples~\cite{goodfellow2014explaining}: adding some imperceptible adversarial perturbations is enough to make them fail. This phenomenon can be found in different applications~\cite{bose2018adversarial,chen2018robust,li2018robust, kurakin2016adversarial, yang2018realistic, tabacof2016adversarial, metzen2017universal, eykholt2017robust}, including classification, object detection, etc. In this paper, we specifically focus on the adversarial attack of object detectors. 

Existing object detectors can be broadly categorized into two groups: anchor-based or anchor-free detectors. Recent anchor-free object detectors~\cite{zhou2019objects, law2018cornernet, zhou2019bottom, huang2015densebox} achieve competitive performance with traditional anchor-base detectors. Additionally, anchor-free detectors are structurally simpler and more computationally efficient than anchor-based detectors. Anchor-based detectors have dominated object detection due to their superior performance. However, adversarial perturbations to this type of detectors have not been explored.

\begin{figure}[t]
    \centering

    \subfigure[Clean Input]{
    \begin{minipage}[t]{0.45\linewidth}
    \centering
    \includegraphics[width=1.5in]{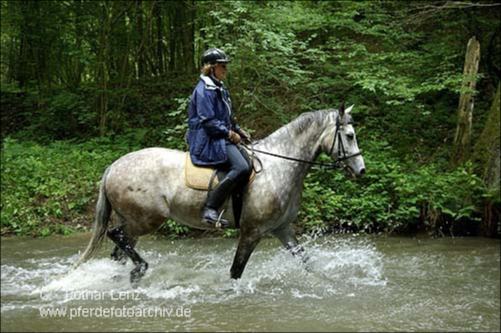}
    \end{minipage}}
    \subfigure[Clean Detection]{
    \begin{minipage}[t]{0.45\linewidth}
    \centering
    \includegraphics[width=1.5in]{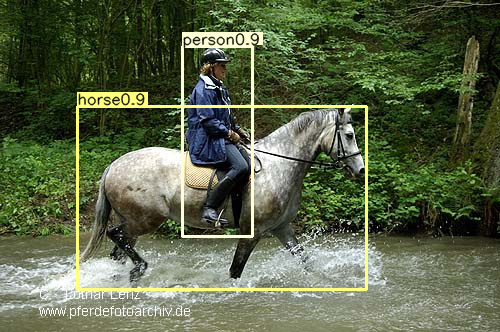}
    \end{minipage}}

    \subfigure[Perturbation of DAG]{
    \begin{minipage}[t]{0.45\linewidth}
    \centering
    \includegraphics[width=1.5in]{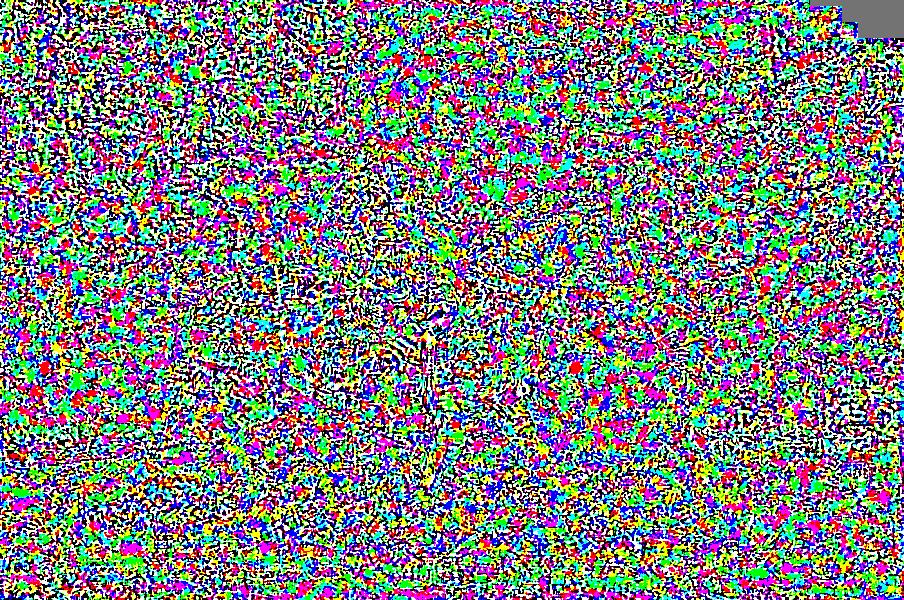}
    \end{minipage}}
    \subfigure[Perturbation of FLA]{
    \begin{minipage}[t]{0.45\linewidth}
    \centering
    \includegraphics[width=1.5in]{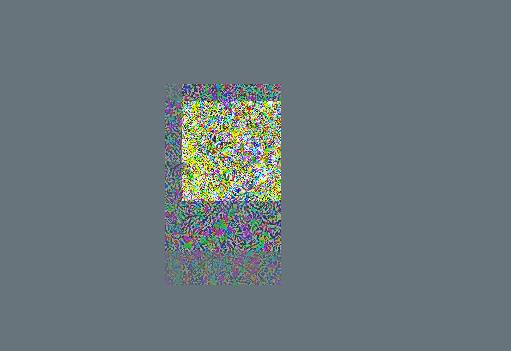}
    \end{minipage}}

    \subfigure[Attack Result of DAG]{
    \begin{minipage}[t]{0.45\linewidth}
    \centering
    \includegraphics[width=1.5in]{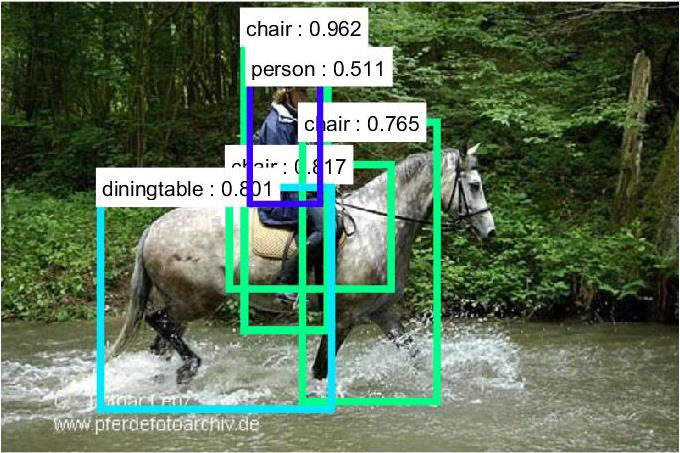}
    \end{minipage}}
    \subfigure[Attack Result of FLA]{
    \begin{minipage}[t]{0.45\linewidth}
    \centering
    \includegraphics[width=1.5in]{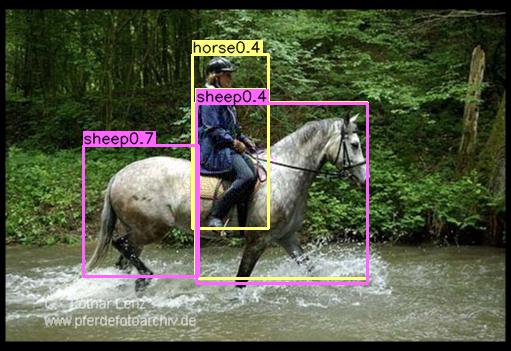}
    \end{minipage}}
    
    \centering
    \caption{Comparison between global (DAG) and local (FLA) attack methods for object detectors. (a) and (b) are the clean input and its detection result on Centernet. In the global attack method, the adversarial examples are generated for all image pixels (c), including the background. Our {Fast Locally Attack (FLA)} generates perturbations only local to the target objects (d). Using this technique, the attack result of FLA (f) is superior to DAG (e). Especially in (d), the FLA only generate small-scale locally perturbation around the center that completes the attack of both human and horse.
    }
    \label{title_fig}
    \vspace{-0.5cm}
\end{figure}

All the existing research focus on the adversarial attack of anchor-based detectors, such as DAG~\cite{xie2017adversarial} and UEA~\cite{wei2019transferable}. However, Existing methods suffer from three major shortcomings: \textbf{1)} Most of the attack methods~\cite{xie2017adversarial, goodfellow2014explaining, moosavi2016deepfool, madry2017towards} generate global perturbations, including the background. However, most of the pixels of these perturbations are useless to fool the detectors. On the contrary, they increase the perceptibility of the perturbations. \textbf{2)} The generated adversarial examples have the poor transferring ability, \emph{e.g.,} the adversarial examples generated by DAG on Faster-RCNN can only attack Faster-RCNN models and the generated examples can hardly be transferred to attack other object detectors. \textbf{3)} They only attack one proposal at one time or relies on training, which is extremely computational consumption. Thus, it is desired to investigate a special attacking scheme for anchor-free detectors that have more local perturbations. 

In this paper, inspired by the fact that the key information lies in or around objects, we propose a new method, \textbf{F}ast \textbf{L}ocally \textbf{A}ttack (FLA), to generate locally adversarial perturbation for anchor-free object detectors. "Fast" denotes that our method generates adversarial examples with lower time consumption than previous work. For this purpose, we focus on high-level semantic information to generate adversarial examples. We attack all objects in each iteration instead of a single object in each iteration, which can reduce the amount of iteration and improve the transferring attack performance of our method. As for "Locally", which means our method generates locally perturbation that only changes little-scale pixels around the detected objects. Locally perturbation can increase the imperceptible of perturbation without reducing attack performance.
Examples in Fig. \ref{title_fig} demonstrates the adversarial examples from our method and compare with DAG's examples. 
Experimental results show that \textbf{FLA} improves the performance over the published state-of-the-art on both white and black box attack tasks.

In comparison with the previous works, the main contributions of this work can be summarized as follows:

\begin{itemize}
\item \textbf{FLA} generates locally perturbations only around target objects, which increases the imperceptibility of the generated perturbations.

\item \textbf{FLA} is efficiently on attacking the SOTA object detectors. It achieves higher white-box attack performance than previous methods with lower computational consumption.

\item The adversarial examples generated by \textbf{FLA} achieve higher black-box attack performance than previous methods. They are not only capable of attacking anchor-free detectors but also can be transferred to attack anchor-based detectors.

\end{itemize}

This paper is organized as follows.  We first discuss the related work in section~\ref{sec: related_work}. Then, we provide the details of \textbf{FLA} in section~\ref{sec: method}. The detailed setup and results of our experiment are described in Section ~\ref{sec: experiment}. Finally, we conclude the paper in section~\ref{sec: conclusion}.

\begin{figure*}[h]
    \centering
    \includegraphics[width=1\linewidth]{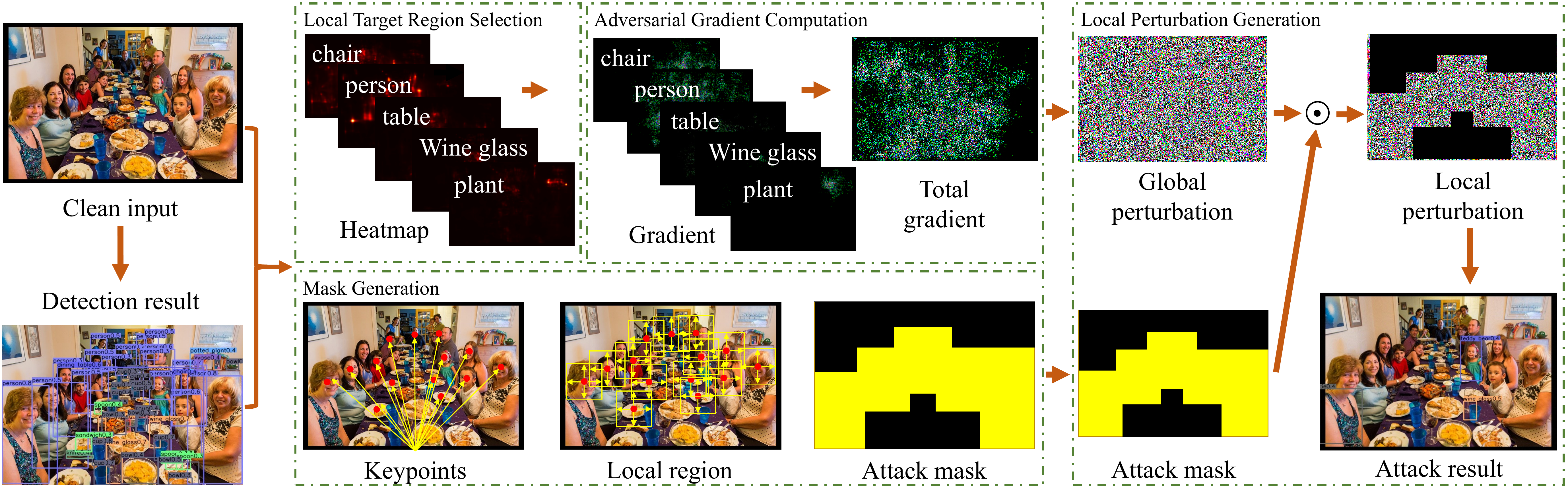}
    \caption{Illustration of Each Iteration of FLA: First of each iteration, we extract heatmap of each detected categories. Second, use the heatmap to compute the adversarial gradient information of each detected categories. Then normalizing the adversarial gradients with $L_{\infty}$ and add up all gradients. After normalizing gradients, add up all gradients to generate globally adversarial perturbation by further applying sign operation on it. Finally, generate locally perturbation by the dot product of global perturbation and mask.}
    \label{gm}
\end{figure*}

\section{Related Work}
\label{sec: related_work}
\subsection{Object Detection}
\noindent

There are some great progress has been made in object detection. With the deep 
convolutional neural network's development, many valuable approaches have been proposed.
One of the most popular object detection categories is the RCNN~\cite{girshick2014rich} family, such as Faster-RCNN~\cite{ren2015faster}. The first process of RCNN's pipeline is generating a large number of proposals which is base on the anchor. Then use a different classifier to classifying the proposals. At last, use post-processing algorithms, such as NMS, to reduce redundancy proposals.

There are also other object detectors that rely on the anchor, like YOLOv2~\cite{redmon2016you}, SSD~\cite{liu2016ssd}. Anchor-base detector has high detected accuracy but also has three shortcomings. Slow, hard to apply a new dataset and more difficult to training. Such as Faster-RCNN.

To solve these shortcomings, some new object detectors have been proposed. Such as CornerNet~\cite{law2018cornernet} and CenterNet~\cite{zhou2019objects}. These new object detectors detect the objects by detecting the keypoints of objects. CornerNet detects the objects by detecting the two corners of the objects. CenterNet relies on find the center points of objects to detect objects. These two methods can complete the training without preset anchor, which we call as the anchor-free detector. These two detectors are not only faster and simpler to training than anchor-base detectors, but they also achieve SOTA detect performance. Both CornerNet and CenterNet can use multiple convolutional neural networks as the backbone network to extract the semantic features of the input image. Then locate the keypoint of the object through these features. Normally, the keypoint include the size and category information of the object. At last, use some post-processing algorithms to remove the redundancy key-point.

\subsection{Adversarial Example for object detection}
\noindent

Goodfellow~\cite{goodfellow2014explaining} first showed the adversarial example problem of the deep neural network. The adversarial example means deliberately generate imperceptibly perturbation to add on the original input dataset. 

The adversarial example is aimed to make the deep neural network output the wrong result.
Almost existing adversarial attack methods are focus on minimizing the $\mathop{L_p}$ norm of the adversarial perturbation. In the most attack methods $\mathop{p = 2}$ or $\mathop{\infty}$ that can generate imperceptible perturbation. 

The most classical attack methods are the FGSM family, such as Fast Gradient Sign Method (FGSM)~\cite{goodfellow2014explaining}, Project Gradient Descent (PGD)~\cite{madry2017towards}. The first of the pipeline of the FGSM is to get the loss value of the deep neural network, then compute the gradient of the input image. At last, use the gradient and the sign function to generate the adversarial perturbation. The principle can be summarized as follows:

\begin{equation}
    \setlength{\abovedisplayskip}{12pt}
    \mathop{x^{\prime} = x + \epsilon\cdot sign(\bigtriangledown_xf(x,y))}
    \label{ep1}
\end{equation}
where $\mathop{f}$ is the classifier, $x$ is the input image for the classifier, 
$\epsilon$ is to constrain the $L_\infty$ of the perturbation.

The difference of the PGD is to add the iterative module to the FGSM. Use many small and accurate perturbation instead of one big perturbation. The PGD achieve higher attack success 
rate and generate smaller $\mathop{L_p}$ norm perturbation.
The principle can be summarized as follows:
\begin{equation}
    \setlength{\abovedisplayskip}{12pt}
    \mathop{x_{t+1} = \Pi_{x+s}(x_{t} + \epsilon\cdot sign(\bigtriangledown_xf(x,y)))}
\end{equation}

There is another attack method can generate lower $\mathop{L_2}$ perturbation than 
FGSM family, Deepfool~\cite{moosavi2016deepfool}. Deepfool uses the generated hyperplane to approximate the decision boundary, 
and compute the lowest Euclidean distance between the input image and the hyperplane iterative. 
Then use the distance to generate the adversarial perturbation. Deepfool achieve state-of-art 
attack performance while has a lower $L_2$ norm of perturbation than the FGSM-base attack method.

The above attack methods are mainly to attack the classifier, there are only a few research for attack object detector. Such as DAG and UEA. Both DAG and UEA attack anchor-base object detectors, 
such as Faster-RCNN and the SSD. Both DAG and UEA generate globally adversarial perturbation.
The main shortcoming of DAG is slow, average consume $10-20s$ to complete ones attack, and weakness 
on transferring attack.
UEA is base on the Generate Adversarial Network~\cite{goodfellow2014generative}, it also achieves high attack performance and better transferring attack performance than DAG. But UEA need retraining to attack new dataset or new detector, 
which is more complex than the optimization-based methods.

\section{Method}
\label{sec: method}

The Fast Locally Attack (FLA) consists of three parts (as shown in Fig. \ref{gm}), i.e., (a) Local Target Region Selection (b) Compute Adversarial Gradient and  (c) Local Perturbation Generation. We first formulate the FLA as a constraint optimization problem in Section (\ref{pd}), then introduce each part in Sections \ref{ltrs}, \ref{agc} and \ref{lpg}, respectively.

\subsection{Problem Definition}
\label{pd}
\noindent


The problem of generating adversarial perturbation for object detection can be formulated to the following constraint optimization problem:

\begin{equation}
    \begin{aligned}
        \mathop{minimize} \limits_{r} \quad & \Vert{r} \Vert_{p} \\
        subject \quad & \widehat{t}(x+r) \cap \widehat{t}(x) = \mathcal{\emptyset} \\
        & min \leq x+r \leq max
    \end{aligned}
\label{pf}
\end{equation}
where ${x}$ is the origin input image, ${r}$ is the adversarial perturbation. $\widehat{t}(x)$ denotes the object set that detected 
by the object detector. $\mathop{max}$, $\mathop{min}$  $\mathop{\in}\mathbb{R}^n$ denote the maximum and minimum pixel intensity to constraint the pixel of the $\mathop{x+r}$.

There are two ways to satisfy that the intersection of $\widehat{t}(x+r)$ and $\widehat{t}(x)$ is empty. 
The first one is attacking each object of $\widehat{t}(x)$ individually, the second one is attacking the whole image to make all object categories incorrect. We found the second way is more efficient, thus constraint in the Eq. (\ref{pf}) can be reformatted as follows:

\begin{equation}
    \begin{aligned}
        \forall t_n \in \widehat{t}(x), f(x+r, t_n) \not= f(x, t_n)
    \end{aligned}
    \label{ep4}
\end{equation}
where $t_n$ denotes the $n$-th object of the object set $\widehat{t}(x)$ that detected by the detector, and the $\mathop{f(x, t_n)}$ denotes the category of the object $\mathop{t_n}$.

\subsection{Local Target Region Selection}
\label{ltrs}
\noindent

We select the local target region using an anchor-free objection method, i.e. CenterNet \cite{zhou2019objects}.
In the CenterNet, the detector removes the classifier module and use the keypoint heatmap $\mathop{\widehat{Y}\in[0, 1]^{\frac{W}{R}\times\frac{H}{R}\times C}}$ to predict the category of the object directly. Where $W$ and $H$ represents the width and height of the heatmap. $C$ means the number of channels. $R$ represents the multiple of downsampling from image to heatmap.

\fussy
The CenterNet use several different CNN networks as the backbone to construct the complete object detector, and output $\mathop{\widehat{Y}}$ of the input image $\mathop{x}$. The $\mathop{\widehat{Y}_{w,h,c} = 1}$ indicates the point $\mathop{\widehat{Y}_{w,h}}$ is a detected keypoint which belonging category $c$, in contrast, the $\mathop{\widehat{Y}_{w,h,c} = 0}$ indicates the point do not belonging $c$. where $w$ and $h$ denotes the abscissa and ordinate of the point. The CenterNet is base on the keypoints detection which regards the center point of the object as the keypoint. Each detected keypoint $\mathop{\widehat{Y}_{w,h,c}}$ denotes the center point of the object. The keypoints include the information of the object's category, it also includes the scale and the offset of the detection box of the object.

Due to the CenterNet relies on the detected keypoints, the attack method can directly attack the detected keypoints to fail the detector. The constraint in Eq. (\ref{ep4}) can be written as the following functions.

\begin{equation}
    \begin{aligned}
        & \widehat{Y} = Centernet(x) \\
        & P = \{p_n = \widehat{Y}_{w,h}~|~\widehat{Y}_{w,h,c} = 1,\widehat{Y}_{w,h,c} \in \widehat{Y}\}\\
        & \forall n, f(x+r, p_n) \not= f(x, p_n), p_n \in P
    \end{aligned}
    \label{ep5}
\end{equation}
where $\mathop{p_n}$ denotes the $n$-th detected keypoints, all detected keypoints construct the target point set $\mathop{P}$.

After attacked all detected keypoints, the CenterNet should fail to detect any object on the adversarial example. But we find it still detects some same object after all the keypoints are attacked. We check the heatmap where all the keypoints are changed to the incorrect category, but the neighbor points around the attacked keypoints are also modified. Some neighbor background points' confidence level is increased that makes them belong to the correct category. Due to the new neighbor keypoints location is near to the old keypoints, the newly detected object is in the same category as the old one and the position and the size of the detected bounding box just change little. These two problems make the CenterNet capable to detect the correct object on the adversarial example. To solve those two problems, we can directly add the neighbor keypoints into the target point set. Then the Eq. (\ref{ep5}) is further extended to the following function.

\begin{equation}
    \begin{aligned}
        & \widehat{Y} = Centernet(x) \\
        & P = \{p_n = \widehat{Y}_{w,h}~|~\widehat{Y}_{w,h,c} = 1,\widehat{Y}_{w,h,c} \in \widehat{Y}\}\\
        & P_{neighbor} = \{p_k~|~p_k \in N(p_n), p_n \in P \}\\
        & P = P \cup P_{neighbor} \\
        & \forall n, f(x+r, p_n) \not= f(x, p_n), p_n \in P
    \end{aligned}
\end{equation}
where $\mathop{N(p_n)}$ indicates the point set which constructed by the neighbor points around the detected keypoint $p_n$.


After selecting $P$ composed of detected points and neighbor points, we divide $P$ into different $P_{j}$ according to different categories $j$ which has detected object.

\begin{equation}
    \begin{aligned}
        & P_{j} = {\{p_n \ \vert \ f(x,p_n) = j, p_n \in P\}}
\end{aligned}
\end{equation}

We generate the adversarial gradient on each $P_{j}$, the process of computing adversarial gradient is summarized in the next section.

\renewcommand{\algorithmicrequire}{\textbf{Input:}}
\renewcommand{\algorithmicensure}{\textbf{Output:}}
\begin{algorithm}[t]
    \setstretch{1.3}
    \caption{Fast Locally Attack (FLA)}
    \label{fla}
    \begin{algorithmic}

        \Require
        image $\mathop{x}$, target points set $\mathop{P}$, number of category $\mathop{C}$ \\
        attack radius $\mathop{R^*}$
        \Ensure
        perturbation $\mathop{r}$

        \State{Initialize: $\mathop{x^{0} \leftarrow x, i \leftarrow 0, j \leftarrow 0, P_{0} \leftarrow P, r_{adv} \leftarrow 0}$}

        \While{$\mathop{P_{i} \cap P \not= \emptyset}$ and $\mathop{i \textless M_D}$}

            \State$\mathop{r_{i} \leftarrow 0, j \leftarrow 0}$
            \State$\mathop{mask_i \leftarrow GenerateMask(x, P_{i}, R^*)}$

            \While{$\mathop{j \textless C}$}

                \State$\mathop{P_{j} = \{p_n \ |\  f(x,p_n) = j, p_n \in P\}}$

                \If{$\mathop{P_{j} \not= \emptyset}$}
                
                    \State$\mathop{loss_{sum} \leftarrow \sum_{p_n\in P_{j}}CrossEntropy(x^{(i)}, p_n)}$
                    \State$\mathop{r_j = \bigtriangledown_{x^{(i)}}loss_{sum}}$
                    \State$\mathop{r_j' = \frac{r_j}{\Vert r_j\Vert_{\infty}}}$
                    \State$\mathop{r_{i} \leftarrow r_{i} + r_j'}$

                \EndIf
                
                \State$\mathop{j \leftarrow j + 1}$

            \EndWhile

            \State$\mathop{x^{i+1} \leftarrow x^{i} + \frac{\epsilon'}{M_D}\cdot sign(r_{i})\cdot mask_i}$
            \State$\mathop{P_{i+1} \leftarrow RefreshPoints\ (x^{i+1}, P_{i})}$
            \State$\mathop{i \leftarrow i + 1}$
  
        \EndWhile

        \State{return $\mathop{r = x^{i} - x^{0}}$}
        
    \end{algorithmic}
\end{algorithm}

\subsection{Adversarial Gradient Computation}
\label{agc}
\noindent

Our method is base on the PGD \cite{madry2017towards}, which generates perturbation iteratively. In each iteration, the perturbation is generated from the adversarial gradient. Unlike DAG, which only computes gradient on a single object and generates perturbation for a single object in each iteration, our method computes gradient on each $P_{target}$ and add up all the gradients to generate perturbation that can reduce the time consumption and increase the transferring attack performance.

As shown in Algorithm. \ref{fla}, during each iteration, we first generate target points set $P_{j}$ for each detected category $j$ which has detected objects.
Then, we compute $loss_{sum}$ of $P_j$ of each detected category $j$ and compute adversarial gradient information 
$r_j$ for each category $j$. Each $r_j$ is normalized by $L_\infty$ and obtain adversarial gradient information $r_j'$. After all, we add up all $r_j'$ to obtain total gradient $r_i$. (An example is shown in Fig. (\ref{gm}).) After obtaining $r_i$, we generate local perturbation in the next section.

\subsection{Local Perturbation Generation}
\label{lpg}
\noindent

To generate locally perturbation, we use attack mask $mask_i$ to keep perturbation that around detected objects and remove perturbation in the background. The attack mask $mask_i$ is generated from the $P_{j}$ by \textit{GenerateMask} step which is demonstrated in Fig. \ref{gm}. At first, we generate a zero matrix $mask$ that has the same size as the input image. We relocated all points $p_i$ of $P_{j}$ on the input image. Then get the location of each $p_i$ and set the same location points of $mask$ as $1$. After relocate all $p_i$, then set all points in the box with the size of attack radius $R^*$ and centered on $p_i$'s location of $mask$ as $1$. After that, we obtain a locally attack mask. In the final of each attack, we use $mask$ and global perturbation to obtain locally perturbation. Different $R^*$ will lead to different attack performance, we will quantitative analysis of this relation in Section \ref{evr}. 

In Algorithm. \ref{fla}. We generate globally perturbation by applying $sign$ operation to the $r_i$. After obtaining global perturbation we generate locally perturbation by the dot product of global perturbation and $mask$. After generating perturbation, we refresh the $P$ by $RefreshPoints$, which remove the points $p_n$ which has been attacked successfully. 


Normally, the DAG needs $\mathop{150-200}$ iterations to generate perturbation and the FLA only needs $\mathop{10-50}$ iterations.

\begin{table*}[t]
    \begin{center}
    \begin{tabular}{|c|c|c|c|c|c|c|}
        \hline
        Method  &Network        & Dataset   & mAP(Clean)    & mAP(Attack)   & ASR           &Time(s)    \\ \hline
        DAG~\cite{xie2017adversarial}     &Faster-RCNN    & PascalVOC & 0.70          & 0.05          & 0.92          &10.0       \\ \hline
        UEA~\cite{wei2019transferable}     &Faster-RCNN    & PascalVOC & 0.70          & 0.05          & 0.92          &---        \\ \hline
        FLA     &Resdcn18       & PascalVOC & 0.67          & 0.07          & 0.90          &\textbf{0.8}\\ \hline
        FLA     &DLA34          & PascalVOC & 0.77          & 0.06          & 0.93          &1.1        \\ \hline
        FLA     &Resdcn18       & MS-COCO      & 0.29          & 0.006         & 0.98          &2.7        \\ \hline
        FLA     &DLA34-1x       & MS-COCO      & 0.38          & 0.008         & \textbf{0.98} &4.7        \\ \hline
    \end{tabular}
    \end{center}
    
    \caption{Results of White-Box Attack (measured by mAP, \%). In the Table, clean means the mAP obtained from 
    the clear input. Attack denotes the mAP obtained from the adversarial example. In the Time column, we show the average attack time of the attack method.}
    \label{tab1}
\end{table*}

\begin{table*}[t]
    \begin{center}
        \begin{tabular}{|c|c|c|c|c|c|c|c|c|c|c|}
        \hline
        \multirow{2}{*}{Network} & \multicolumn{2}{l|}{Resdcn18} & \multicolumn{2}{l|}{Resdcn101} & \multicolumn{2}{l|}{DLA34-1x} & \multicolumn{2}{l|}{DLA34-2x} & \multicolumn{2}{l|}{CornerNet} \\ \cline{2-11} 
                                 & mAP           & ATR           & mAP            & ATR           & mAP           & ATR           & mAP           & ATR           & mAP            & ATR           \\ \hline
        Clean                    & 0.29          & ---           & 0.36           & ---           & 0.38          & ---           & 0.39          & ---           & 0.43           & ---           \\ \hline
        DLA34-1x                 & 0.10          & 0.86          & 0.12           & 0.87          & \textbf{0.09} & 1.00          & 0.11          & 0.94          & 0.13           & 0.92          \\ \hline
        DLA34-2x                 & \textbf{0.10} & \textbf{0.88} & \textbf{0.12}  & \textbf{0.90} & 0.11          & 0.96          & \textbf{0.10} & 1.00          & \textbf{0.13}  & \textbf{0.94} \\ \hline
        \end{tabular}
    \end{center}

    \caption{Test black-box attack performance on the COCO dataset. The first column means which network that adversarial examples generate from.
    The first row means which network that been attacked in the black-box attack.}
    \label{ba_coco}
\end{table*}

\begin{table*}[t]
    \begin{center}
        \begin{tabular}{|c|c|c|c|c|c|c|c|c|c|c|}
            \hline
            \multirow{2}{*}{Network} & \multicolumn{2}{l|}{Resdcn18} & \multicolumn{2}{l|}{DLA34} & \multicolumn{2}{l|}{Resdcn101} & \multicolumn{2}{l|}{Faster-RCNN} & \multicolumn{2}{l|}{SSD300} \\ \cline{2-11} 
                                     & mAP           & ATR           & mAP          & ATR         & mAP            & ATR           & mAP             & ATR            & mAP          & ATR          \\ \hline
            Clean                    & 0.67          & None          & 0.77         & None        & 0.76           & None          & 0.71            & None           & 0.77         & None         \\ \hline
            DAG~\cite{xie2017adversarial}                      & 0.65          & 0.19          & 0.75         & 0.16        & 0.74           & 0.16          & 0.60            & 1.00           & 0.76         & 0.08         \\ \hline
            DLA34-384                & 0.50          & 0.30          & 0.1          & 1.00        & 0.62           & 0.22          & 0.53            & \textbf{0.35}  & 0.67         & 0.15         \\ \hline
            DLA34-512                & \textbf{0.48} & \textbf{0.32} & \textbf{0.07}& 1.00        & \textbf{0.60}  & \textbf{0.24} & \textbf{0.51}   & 0.37           & \textbf{0.66}& \textbf{0.16}\\ \hline
        \end{tabular}
    \end{center}

    \caption{Test black-box attack performance on the Pascal dataset. The first column means which network that adversarial examples generate from.
    The first row means which network that been attacked in the black-box attack. The row 'DAG' denotes the black-box attack result of DAG. The row 
    'DLA34-384' and 'DLA34-512' denotes the black-box attack result of FLA on DLA34 backbone centernet with different input scale.}
    \label{ba_pascal}
\end{table*}

\begin{figure*}[t]
   \centering
   \includegraphics[width=1\linewidth]{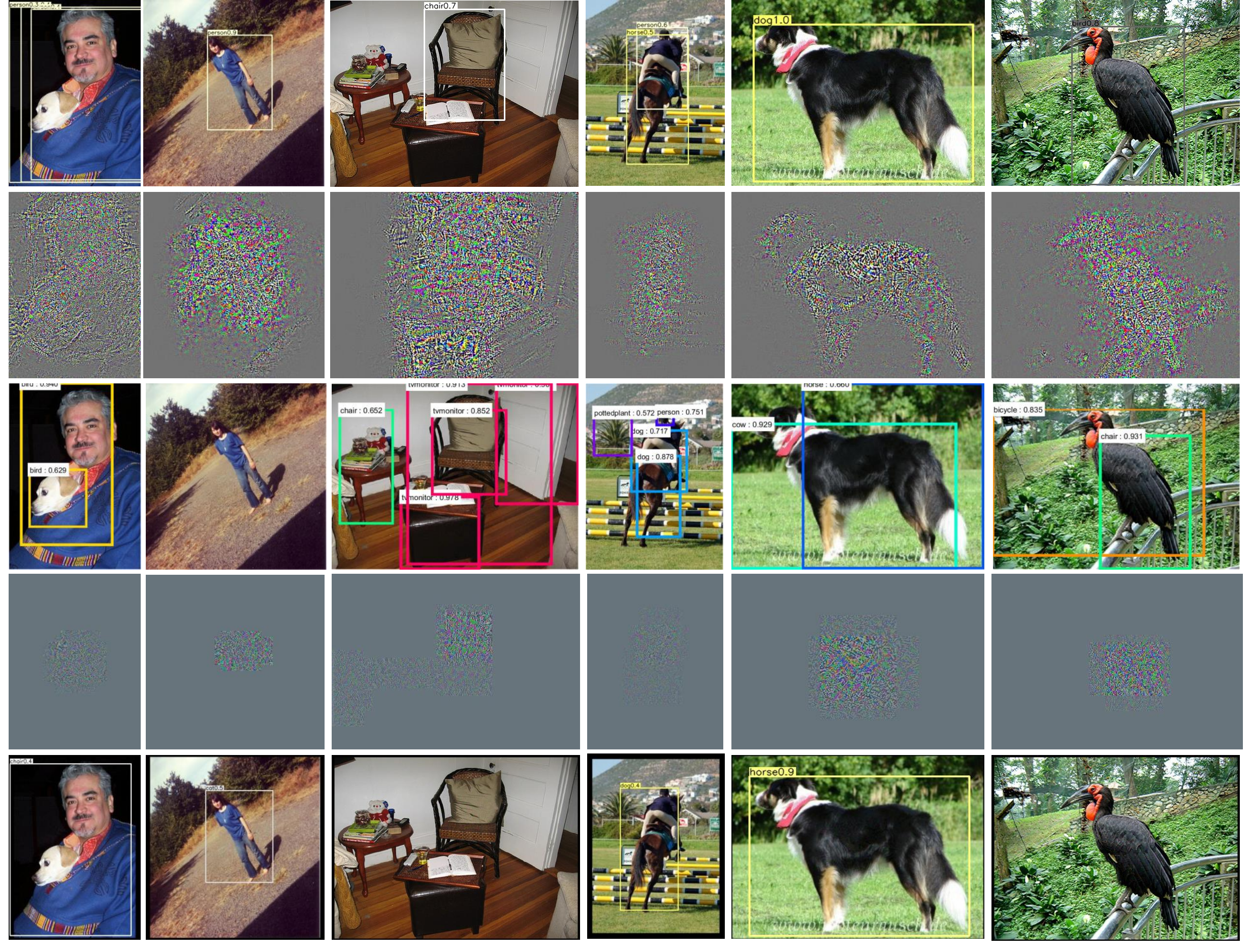}
   \caption{Each column is an example. \textbf{Row 1:} Detection results of clean inputs on CenterNet. \textbf{Row 2$\&$3:} DAG perturbations and DAG attacked results on Faster-RCNN. \textbf{Row 4$\&$5:} FLA perturbations and FLA attacked results on CenterNet. Note that in \textbf{Row 4}, from left to right, the percentage of the changed pixels for each FLA perturbations are: 17\% 8\%, 26\%, 25\%, 26\%, 14\%. We can see that the perturbations of FLA are smaller than the DAG. To better show the perturbation, we have multiplied the intensity of all perturbation images by 10.}
   \label{qualitative_1}
\end{figure*}

\section{Experiment}\label{sec: experiment}
In this section, we first introduce the detailed setup of the experiment. Then, we report both the white-box and black-box attack results. Finally, we evaluate the perceptibility of the generated adversarial examples and the attack radius $R^*$.
\subsection{Experimental Details}

\textbf{1)~Attacked Object Detectors}: All adversarial examples are generated on CenterNet with two different backbones: ResNet-18~\cite{he2016deep} and DLA-34~\cite{yu2018deep}. CenterNet with backbone ResNet-18 is highly efficient. In contrast, CenterNet with backbone DLA-34 is more computationally intensive but more accurate.


\textbf{2)~Dataset}: 
The above two networks are trained on the training set of PascalVOC~\cite{everingham2015pascal} and MS-COCO~\cite{lin2014microsoft}. The training set of PascalVOC includes the trainval sets of PascalVOC-2007 and PascalVOC-2012. In this paper, both the white-box and black-box attack performance are reported on the testing set of PascalVOC and MS-COCO.

\textbf{3)~Metrics:} We compare the white-box attack performance with the DAG and the UEA. It is evaluated by computing the decreased percentage of mean average precision ($mAP$), which is referred to as \textbf{Attack Success Ratio(ASR)} in this paper:
\begin{equation}
    \begin{aligned}
        ASR = 1-\frac{mAP_{attack}}{mAP_{clean}}
    \end{aligned}
\end{equation}
where $mAP_{attack}$ denotes the $mAP$ of the targeted object detector on adversarial examples. $mAP_{clean}$ denotes the $mAP$ of clean input.
Higher $ASR$ means better white-box attack performance.

The black-box attack signifies the transferability of the generated adversarial examples to other object detectors. In this paper, black-box adversarial examples are generated on Centernet with DLA-34~\cite{yu2018deep} backbone and tested on Centernet with different backbones (Resdcn18 and Resdcn101). We also test these adversarial examples on other object detectors, including anchor-free (CornerNet) and anchor-based detectors (Faster-RCNN and SSD300). In this paper, the performance of the black-box attack is evaluated by the ASR ratio between the targeted detector and the original detector on which the adversarial examples are generated. It's referred to as \textbf{Attack Transfer Ratio (ATR)} in this paper:
\begin{equation}
    \begin{aligned}
        ATR = \frac{ASR_{target}}{ASR_{origin}}
    \end{aligned}
\end{equation}
where $\mathop{ASR_{target}}$ represents the $\mathop{ASR}$ of the targeted detector and $\mathop{ASR_{origin}}$ denotes the $\mathop{ASR}$ of the detector on which the adversarial examples are generated. Higher ATR denotes better transferability. 



\textbf{4)~Perceptibility Metric:} The adversarial perturbation's perceptibility is quantified by its $L_p$ norms. Specifically, $P_{L_2}$ and $P_{L_0}$ are used, which are defined as follows.

\textbf{i) $P_{L_2}$:} $L_2$ norm of the perturbation. A lower $L_2$ value usually signifies that the perturbation is more imperceptible for the human. Formally, 
\begin{equation}
    P_{L_2} = \sqrt{\frac{1}{k}\sum{r_k^2}}
\end{equation}
where the $\mathop{k}$ is the number of the pixels. We also normalized the $\mathop{P_{L_2}}$ in 
$\mathop{[0,1]}$. 

\textbf{ii) $P_{L_0}$:} $L_0$ norm of the perturbation. A lower $L_0$ value means that less less images images are changed during the attack. We compute $P_{L_0}$ by measuing the proportion of changed pixels. The whole experiment is conducted with a Intel Core i7-7700k CPU and an Nvidia GeForce GTX-1080ti GPU.



\begin{figure*}[t]
    \centering
    \includegraphics[width=1\linewidth]{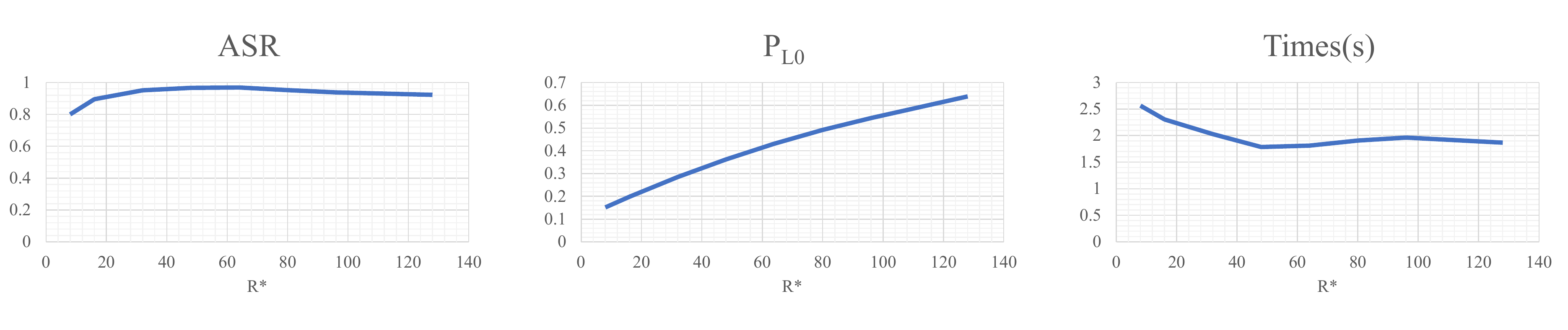}
    \vspace{-1.0cm}
    \caption{Correlation between the attack radius $\mathop{R^*}$ and the attack performance, time consumption, and the perceptibility of adversarial examples. The $ASR$ of FLA correlates positively with $R^*$ when $R^*$ is less 16. However, $ASR$ is stable or slightly decreased afterwards. $P_{L_0}$ of the perturbation correlates positively with $R^*$. This is because higher $R^*$ means bigger attack masks. The mean attack time of FLA correlates negatively with $R^*$ when $R^*$ is lower than 48. However, the mean attack time of FLA become stable afterwards.}
    \label{ar}
\end{figure*}

\subsection{White-Box Attack Results}
\noindent
In this subsection, we show the white-box attack result on PascalVOC and MS-COCO. The overall attack results are shown in Table \ref{tab1}. It is obvious that the mAPs of different Centernet have dropped dramatically after adversarial attack. In PascalVOC, the ASR of FLA is 0.90 and 0.93 respectively when the backbone is Resdcn18 and DLA-34, outperforming DAG and UEA. Besides, FLA is almost ten times faster than DAG. Regarding UEA, we set the attack time to N/A, as they don't provide the source code. Note that UEA requires additional training time. In MS-COCO, the FLA achieves $\mathop{0.98}$ ASR on Resdcn18 and DLA34. On Resdcn18, the average attack time required by FLA is 2.7s. On DLA34, the average attack time required by FLA is 4.7s.




\subsection{Black-Box Attack Results}
\noindent

We report the black-box attack results in this subsection. The black-box attack measures the transferability of adversarial examples. All adversarial examples generated for black-box attack experiment are generated on Centernet with backbone DLA34-1x and DLA34-2x. 

At first in the experiment, we use the FLA to generate the adversarial example on the CenterNet and 
save the adversarial example in a common image format, JPG. Then reload the saved adversarial example 
to compute the mAP. We want to simulate a real transferring attack scenario, so we are abandoning the use of 
lossless photo formats and use the JPG format to save. Most of transferring test use the lossless float matrix 
to test, but most of normal image input is $\mathop{8-bit}$ int matrix. So we save the adversarial example 
in JPG format, this process will better simulate a real transferring attack scenario.
Compare with directly compute mAP with lossless adversarial example, save the adversarial
example in JPG will lose a small amount of aggressiveness~\cite{dziugaite2016study}, because save in JPG will lose some details of the adversarial example. But in this way we can further guarantee the attack ability of our method.
We test the transferability of adversarial examples by generated on one model and compute mAP on other models.

\begin{table}[t]
    \begin{center}
    \begin{tabular}{|c|c|c|c|}
        \hline
        Network         & Dataset   & $\mathop{P_{L_2}}$             & $\mathop{P_{L_0}}$ \\ \hline
        DAG~\cite{xie2017adversarial}             & PascalVOC & $\mathop{3\times 10^{-3}}$    & $\mathop{>99\%}$ \\ \hline
        Resdcn18        & PascalVOC & $\mathop{5.9\times 10^{-3}}$  & $\mathop{30\%}$ \\ \hline
        DLA34           & PascalVOC & $\mathop{6.1\times 10^{-3}}$  & $\mathop{32\%}$ \\ \hline
        Resdcn18        & MS-COCO      & $\mathop{6.0\times 10^{-3}}$  & $\mathop{38\%}$ \\ \hline
        DLA34           & MS-COCO      & $\mathop{6.0\times 10^{-3}}$  & $\mathop{36\%}$ \\ \hline
    \end{tabular}
    \end{center}
    
    \caption{Evaluation of the perceptibility of the generated adversarial examples. Higher $P_{L_2}$ and $P_{L_2}$ values generally mean that the generated adversarial examples are more perceptible to human eyes.}
    \label{tab2}
    \vspace{-0.5cm}
\end{table}

As the pervious experiment, we evaluate the black-box attack performance of the adversarial example on PascalVOC and MS-COCO. On the PascalVOC, we generate adversarial example on DLA34-1x and DLA34-2x backbone centernet. As compare, we generate adversarial example on Faster-RCNN by DAG and also save as JPG to transferring to other detector. On the COCO, we also generate adversarial example on DLA34-1x and DLA34-2x backbone centernet. and transferring to other backbone centernet and CornerNet.

The results of black-box attack are summarize in the Table~\ref{ba_coco} and Table~\ref{ba_pascal}.
As the shown on Table~\ref{ba_pascal}. On the PascalVOC, adversarial example that generated by our method has obviously higher 
black-box attack performance than the DAG. The adversarial example that generated by DAG loss its aggressiveness after 
JPG compression, hard to attack the faster-rcnn. The adversarial example that generated by FLA keep its aggressiveness 
after JPG compression, and achieve higher ATR than the DAG. The adversarial example that generated by FLA is also valid to 
transferring attacking anchor-base detector, Faster-RCNN and SSD300. Meanwhile, the adversarial example generated by 
DAG is invalid to attack anchor-free detector.

On the Table~\ref{ba_coco}, compare with test on PascalVOC, FLA achieve higher ATR on COCO. The average ATR is over 90\%. The 
adversarial example that generated by FLA is valid to different backbone centernet. It is also achieve high black-box attack 
performance when transferring to CornerNet.

\subsection{Evaluation of Perceptibility}
The $\mathop{P_{L_2}}$ and $\mathop{P_{L_0}}$ values of the generated adversarial examples by FLA are summarized in Table~\ref{tab2}. Higher $P_{L_2}$ and $P_{L_2}$ values generally mean that the generated adversarial examples are more perceptible to human eyes. Although $\mathop{P_{L_2}}$ of our method is slighly higher than DAG, the perturbation is almost  imperceptible to human eyes (see Fig.~\ref{qualitative_1}). 
The $\mathop{P_{L_0}}$ value of FLA is significantly lower than DAG, meaning that the generated perturbations of FLA are much locally constrained than DAG.

\subsection{Evaluation of Attack Radius $R^*$}
\label{evr}
$\mathop{R^*}$ correlates with the attack performance, time consumption, and the perceptibility of adversarial examples. We summarize the relation between $\mathop{R^*}$ and adversarial perturbation in Fig. \ref{ar}. In this experiment, all adversarial examples are generated by attacking centernet with backbone Resdcn18 on MS-COCO. Based on Fig.~\ref{ar}, we can draw three conclusions. First, the $ASR$ of FLA correlates positively with $R^*$ when $R^*$ is less 16. However, $ASR$ is stable or slightly decreased afterwards. Second, $P_{L_0}$ of the perturbation correlates positively with $R^*$. This is because higher $R^*$ means bigger attack masks. Finally, the mean attack time of FLA correlates negatively with $R^*$ when $R^*$ is lower than 48. However, the mean attack time of FLA become stable afterwards.


\section{Conclusion}
\label{sec: conclusion}

In this paper, we propose {Fast Locally Attack} to generate transferable adversarial example for attacking SOTA anchor-free object detectors. Our method leverages higher-level semantic information to generate locally adversarial perturbation. FLA computes adversarial gradient information for all detected categories and generate locally perturbation, which can improve the attack performance of adversarial example. FLA also achieved SOTA white-box attack performance for attacking Centernet, while being dozens of times faster than DAG. Additionally, it only changes $30\%-40\%$ image pixels during the attack process. Finally, our method achieved better black-box attack performance and are more robust to JPEG compression. 

\section*{Acknowledgment}

This work was supported by the Sichuan Science and Technology Program under Grant 2019YFG0399.

\bibliographystyle{plain}
\bibliography{main}

\vspace{12pt}

\end{document}